\documentclass[twoside,11pt]{article}
\usepackage{pgm}

\usepackage[utf8]{inputenc}
\inputencoding{utf8}

\usepackage{hyperref,color,soul,booktabs}
\usepackage{pdflscape}
\usepackage{natbib}
\usepackage{hyperref,color,soul,booktabs}
\usepackage{multirow}
\usepackage{float}
\usepackage{amsmath}
\usepackage{makecell}
\usepackage{graphicx}
\setulcolor{blue}
\newcommand{\BibTeX}{\textsc{B\kern-0.1emi\kern-0.017emb}\kern-0.15em\TeX}
\newcommand{\mysubscript}[1]{\raisebox{-0.34ex}{\scriptsize#1}}

\ShortHeadings{PGM 2022: Discovery and density estimation of latent confounders}{Chobtham and Constantinou}

\begin{document}

\title{Discovery and density estimation of latent confounders in Bayesian networks with evidence lower bound}

\author{\Name{Kiattikun Chobtham} \Email{k.chobtham@qmul.ac.uk}\and
 \Name{Anthony C. Constantinou} \Email{a.constantinou@qmul.ac.uk}\\
 \addr \href{http://bayesian-ai.eecs.qmul.ac.uk/}{Bayesian Artificial Intelligence research lab}, School of Electronic Engineering and Computer Science, Queen Mary University of London, London, UK, E1 4NS.}

\maketitle

\begin{abstract}
Discovering and parameterising latent confounders represent important and challenging problems in causal structure learning and density estimation respectively. In this paper, we focus on both discovering and learning the distribution of latent confounders. This task requires solutions that come from different areas of statistics and machine learning. We combine elements of variational Bayesian methods, expectation-maximisation, hill-climbing search, and structure learning under the assumption of causal insufficiency. We propose two learning strategies; one that maximises model selection accuracy, and another that improves computational efficiency in exchange for minor reductions in accuracy. The former strategy is suitable for small networks and the latter for moderate size networks. Both learning strategies perform well relative to existing solutions.
\end{abstract}
\section{Introduction}
Methods from both machine learning and statistical sciences have contributed to the advancements in probabilistic graphical models, and specifically in learning Bayesian Networks (BNs). The structure of a BN is typically represented by a Direct Acyclic Graph (DAG), containing nodes that represent variables and edges that represent conditional relationships. The process of learning BNs from data is separated into two tasks. An unsupervised structure learning approach is first used to determine the graph of the BN, followed by parameter estimation of the conditional distributions given the learnt structure.
\par The task of structure learning is known to be both difficult and computationally expensive, and it is NP-hard even for small networks containing just 10 variables. These challenges are elevated when noise is present in the data, or when the data are incomplete \citep{Largescale}. One such example is when the input data do not capture all the variables; often referred to as the problem of learning under the assumption of causal insufficiency \citep{RePEc:mtp:titles:0262194406}. A special case of a latent variable is the latent confounder which represents a missing common cause of two or more observed variables, and tends to lead to spurious edges between its children. 
\par Structure learning algorithms that assume causal insufficiency generate ancestral graphs that represent an extension of a DAG. These algorithms generate either a Maximal Ancestral Graph (MAG) that contains bi-directed edges indicating confounding and directed edges indicating causal or ancestral relationships, or a Partial Ancestral Graph (PAG) that represents the Markov equivalence class of a set of MAGs, in the same way a Completed Partial DAG (CPDAG) represents the Markov equivalence class of a set of DAGs. As we later explain in subsection 2.1, a PAG also captures the possible existence of multiple latent confounders. Well-established algorithms that generate PAGs tend to be constraint-based or hybrid, and largely rely on FCI by Spirtes et al. \citeyearpar{RePEc:mtp:titles:0262194406}. Some variants of FCI include the constraint-based RFCI algorithm by Colombo et al. \citeyearpar{colombo}, cFCI by Ramsey et al. \citeyearpar{DBLP:journals/corr/RamseyZS12}, mFCI by Colombo and Maathuis \citeyearpar{JMLR:v15:colombo14a}, and the hybrid GFCI algorithm by Ogarrio et al. \citeyearpar{pmlr-v52-ogarrio16} which combines elements of constraint-based and score-based learning. 
\par In this work, we describe two learning strategies that take a PAG as an input, along with observed data, to determine the existence of latent confounders and learn their underlying latent distributions. We propose two algorithms: one that maximises accuracy and another that balances accuracy with computational complexity. The paper is organised as follows: Section 2 provides preliminary information through past related works, Section 3 describes the two algorithms, Section 4 describes the experimental setup, Section 5 presents the results, and we provide our conclusions and directions for future work in Section 6.
\section{Preliminaries}
\subsection{Ancestral Graphs}
As discussed in the introduction, a MAG represents a DAG extension that captures information about possible latent confounders. A PAG, which represents a set of Markov equivalent MAGs that entail the same set of Conditional Independencies (CIs) or m-separation criteria, is represented by a tuple $\left(V,E\right)$ where $V$ is the set of observed variables and $E$ is the set of the edges. The edges in a PAG can be: —, $\leftrightarrow$, $\rightarrow$, o$\rightarrow$ or o—o, where — indicates selection variables, $\leftrightarrow$ indicates latent confounders, and $\rightarrow$ indicates that all MAGs in the equivalence class contain this directed edge. The circle edge mark (o) indicates that the endpoint of the edge could be either a tail (–) or an arrowhead ($>$). For example,  o$\rightarrow$ indicates that the corresponding MAGs will contain either $\rightarrow$ or $\leftrightarrow$, whereas o—o indicates that the edge can be $\rightarrow$, $\leftarrow$ or $\leftrightarrow$. Because we do not deal with selection bias in this paper, we will not be considering ancestral graphs that contain undirected edges (—). Both PAGs and MAGs are acyclic and do not assume partially directed cycles when $A$$\leftrightarrow$$B$; but instead assume that either $B$ is an ancestor of $A$, or $A$ is an ancestor of $B$ \citep{richardson}. 
\subsection{Conjugate-exponential family models}
We consider conjugate-exponential family models for discrete data. We assume a Dirichlet prior that serves as a conjugate prior of a multinomial likelihood \citep{bishop:2006:PRML}, whose posterior distribution is also Dirichlet. We use the empirical Bayes method by Gelman et al. \citeyearpar{gelman2003bayesian} to determine the prior parameters from data. For density estimation of latent confounders, we assume a Dirichlet prior $Dir\left(\theta_i|\alpha_{ik}\right)$ where $\alpha_{ik}$ is a hyperparameter set to ‘1’ for uniform distribution, and $\theta_i$ denotes parameters $\sum_{k}\theta_{ik}=1$ where $k$ represents the number of states. Since we perform structure learning and density estimation under causal insufficiency, some variables will not be observed in the data, leading to an incomplete-data marginal likelihood $p(D|G)$ of a DAG $G$.
\subsection{Variational Bayesian Expectation-Maximization (VBEM) algorithm }
Marginalising out the parameters over latent confounders $L_j$ in $p(D|G)$ makes the task of learning prohibitively expensive and intractable. We address this issue by approximating distributions of latent variables using the computationally efficient Variational Bayesian Expectation-Maximization (VBEM) algorithm \citep{Bernardo03thevariational} that enables tractable solutions. The VBEM algorithm combines elements of variational inference \citep{VI} and Expectation-Maximisation (EM; \citealp{EM}). It uses an alternated optimisation technique to find a surrogate distribution $q\left(L,\theta\right)$ from any exponential family $Q$ (e.g., Gaussian, Dirichlet, multinomial) and optimises towards the true distribution $p\left(L,\theta|D,G\right)$. VBEM offers an approximate solution that guarantees to monotonically increase the objective score, and scales better with large data compared to MCMC \citep{MCMC}.
\par The objective of VBEM is to minimise the discrepancy between two distributions $q\left(L,\theta\right)$ and $p\left(L,\theta|D,G\right)$. It uses the reverse Kullback-Leiber (KL) divergence for this task, which is the standard choice for variational inference, defined as follows:
\begin{flushleft}
\,\,\,\,\,\,\,\,\,\,\,\,\,\,\,\,\,\ $KL\left(q\parallel p\right)=\iint{dLd\theta q\left(L,\theta\right)log{\frac{q\left(L,\theta\right)}{p\left(L,\theta|D,G\right)}}\ }$
\end{flushleft}
\begin{flushleft}
\,\,\,\,\,\,\,\,\,\,\,\,\,\,\,\,\,\,\,\,\,\,\,\,\,\,\,\,\,\,\,\,\,\,\,\,\,\,\,\,\,\, $\ \ =\mathbb{E}_q\left[log\frac{q\left(L,\theta\right)}{p\left(L,\theta|D,G\right)}\right]$
\end{flushleft}
\begin{flushleft}
\,\,\,\,\,\,\,\,\,\,\,\,\,\,\,\,\,\,\,\,\,\,\,\,\,\,\,\,\,\,\,\,\,\,\,\,\,\,\,\,\,\, $\ \ =\mathbb{E}_q\left[log{p\left(D|G\right)}\right]-\left\{\mathbb{E}_q\left[log{p\left(L,\theta,D|G\right)}\right]-\mathbb{E}_q\left[log{q\left(L,\theta\right)}\right]\right\}\ \ \ (1)$
\end{flushleft}
Because the incomplete-data marginal likelihood $p\left(D|G\right)$ is intractable to compute, we consider $p\left(D|G\right)$ to be a constant. The aim is to minimise $KL\left(q\parallel p\right)$, which is equivalent to maximising the Evidence Lower Bound (ELBO). Therefore, we can minimise $KL\left(q\parallel p\right)$ without having to know the true distribution $p\left(L,\theta|D,G\right)$ and $p\left(D|G\right)$. We can describe ELBO as the objective function:
\begin{center} 
ELBO$\ =\mathbb{E}_q\left[log{p(L,\theta,D|G}\right)]-\mathbb{E}_q\left[log{q\left(L,\theta\right)}\right]\ \ (2)$
\end{center}
where $q\left(L,\theta\right)$ is assumed to be the factorisation of the free distributions $q_L\left(L\right)$ and $q_\theta\left(\theta\right)$. We maximise ELBO using a function $\mathcal{F}\ $of both $q_L\left(L\right)$ and $q_\theta\left(\theta\right)$ as follows \citep{Beal2006VariationalBL}:
\begin{center} 
ELBO$\ =\mathcal{F}\left(q_L\left(L\right),q_\theta\left(\theta\right)\right)=\iint{dLd\theta q_L\left(L\right)q_\theta\left(\theta\right)\left[logp\left(L,\theta,D|G\right)-log\left({q_L\left(L\right)q}_\theta\left(\theta\right)\right)\right]\ (3)}$
\end{center}
To maximise $\mathcal{F}$, VBEM calculates $q_L\left(L\right)$ and $q_\theta\left(\theta\right)$ while holding the other fixed at iteration $t$. The two steps for each iteration $t$ are:
\\
\par \textbf{1) VB-E step}: estimates the posterior distribution over latent confounders $q_L^{t+1}\left(L\right)=\prod_{i=1}^{\left|L\right|}{q_{L_i}^{t+1}\left(L_i\right)}$ given $q_\theta^t\left(\theta\right)\ $ from the last iteration by taking the functional derivatives in Equation (3) with respect to $q_{L_i}\left(L_i\right)$, where $|L|$ is the number of latent confounders.
\\
\par \textbf{2) VB-M step}: estimates $q_\theta^{t+1}\left(\theta\right)$ given the posterior distribution ${q_L}^{t+1}\left(L\right)$ taken from the VB-E step by taking the functional derivatives in Equation (3) with respect to $q_\theta\left(\theta\right)$.
\\
\par VBEM iterates over the VB-E and VB-M steps until the difference in ELBO becomes smaller than a given threshold, indicating convergence. Since ELBO is not a score-equivalent function, it generates different values for graphs that belong to the same Markov equivalence class. A revised version called p-ELBO was proposed by Rodriguez-Sanchez et al. \citeyearpar{9207730} that includes a penalty term to avoid the $\left|L_i\right|!$ equivalent ways of assigning sets of parameters that result in the same distribution (non-identifiability), and it is defined as p-ELBO = ELBO$\ -\sum_{i=1}^{|L|}{log\ \left|L_i\right|!}$; where $\left|L_i\right|$ is the number of states in $L_i$.
\subsection{Past relevant work}
ELBO was used as the objective function of a neural network in Variational Autoencoder (VAE) by Kingma and Welling \citeyearpar{VAE}. VAE for heterogeneous Mixed type data (VAEM) was used by Ma et al. \citeyearpar{VAEM} for density estimation of latent variables in deep generative models. VAE assumes each observed variable has a latent parent, whereas VAEM is an extension of VAE that assumes an additional latent confounder that serves as a parent of all latent variables.
\par The ELBO score was extended to p-ELBO by Rodriguez-Sanchez et al. \citeyearpar{9207730,GLSL}, which was used as the objective score in Constrained Incremental Learner (CIL) and Greedy Latent Structure Learner (GLSL) algorithms. CIL learns a tree-structured BN that assumes any two nodes are connected by one directed path only, whereas GLSL learns a DAG BN. Both algorithms start from an empty graph and perform various search operations including a) add or remove latent variables as parents of observed variables, b) increase the number of states of latent variables, and c) perform edge operations such as add, remove, or reverse edges, aiming to maximise p-ELBO. Searching for latent confounders often means iterating over all pairs of observed variables, which can be computationally expensive. Instead, these algorithms offer a strategy that focuses on a set of pairs of variables that provide the highest Mutual Information (MI). Empirical results show that GLSL outperforms CIL, but at the expense of high computational complexity.
\section{Two new algorithms for learning latent confounders}
This section describes the two learning strategies we have implemented for latent confounder discovery and density estimation. Subsection 3.1 describes how we use existing algorithms to draw a PAG that is then given as an input to the two algorithms we propose, which in turn use the PAG to search for different MAGs and DAGs with parameterised latent confounders. We describe the two algorithms in subsections 3.2 and 3.3 respectively. Both algorithms assume the input data are discrete, and that the latent confounders have no parents but have at least two children. We further assume a Dirichlet prior  $q_\theta\left(\theta\right)$ over all parameters as described in subsection 2.2, and we use p-ELBO as the objective function which is computed using the VBEM algorithm as described in subsection 2.3.
\subsection{Searching for MAGs and DAGs given a PAG input}
The FCI algorithm and some of its variants discussed in the introduction represent the state-of-the-art in recovering ancestral graphs under the assumption of causal insufficiency \citep{survey}. Any of these algorithms can be used to draw a PAG that can be given as input to the two algorithms described in subsections 3.2 and 3.3. A set of Markov equivalent MAGs can be then derived from the input PAG. However, because the number of possible latent confounders that can be explored for a given MAG is generally intractable, we shall assume the minimum number of latent confounders that satisfy the m-separation criteria. Since the computational cost of working with multiple latent confounders is high, it becomes necessary that we introduce Assumption 1, as described below.
\\
\\
\textbf{Assumption 1}: The optimal number of latent confounders is the minimum number of latent confounders that retain the CIs of a given MAG.

\includegraphics[scale=0.5]{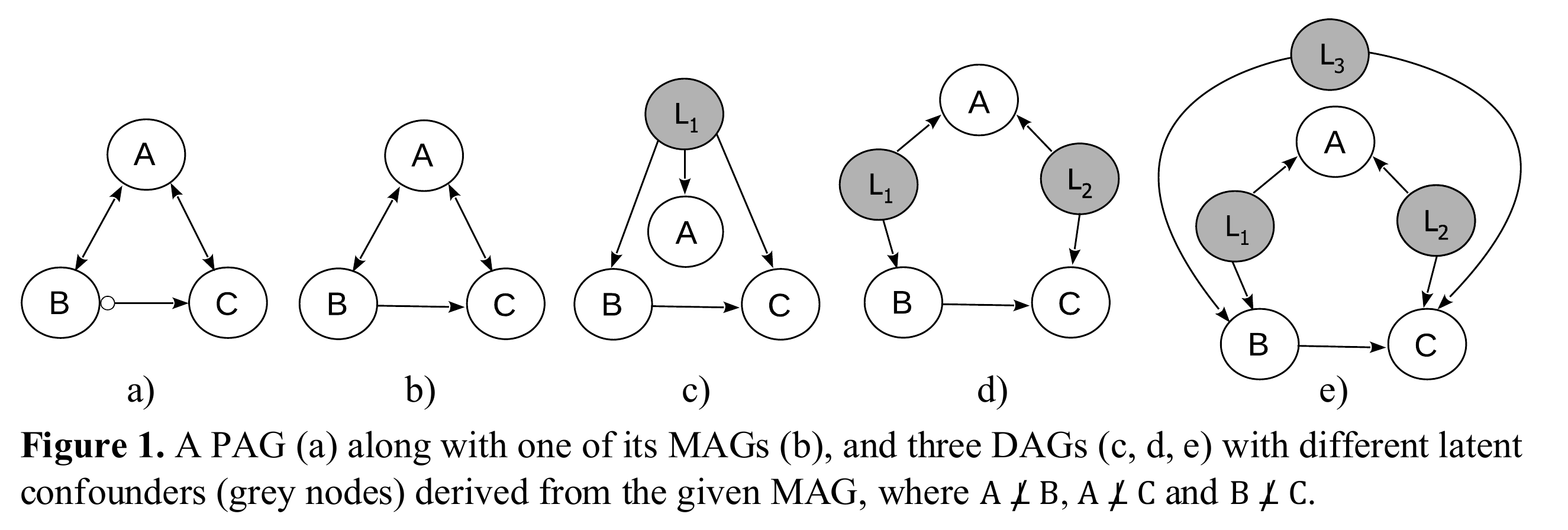}
\par Figure 1 presents a simple PAG that contains two bi-directed edges, along with a MAG and three DAGs that satisfy the CI statements of the PAG. Converting a MAG into possible DAGs implies that each DAG retains the CIs of that MAG by reducing the criteria of m-separation to d-separation. In this example, the DAG that contains the minimum number of latent confounders, with reference to the MAG in Figure 1b, is shown in Figure 1c. The DAGs in Figures 1d and 1e contain a higher number of latent confounders than the minimum required to satisfy all the CIs of the given MAG. Because the algorithms we describe in subsection 3.2 and 3.3 rely on \textbf{Assumption 1}, they will never explore DAGs that contain a higher number of latent confounders than the minimum required, and would not visit DAGs such as those shown in Figures 1d and 1e.
\subsection{Alg 1: Incremental Latent Confounder search with VBEM (ILC-V)}
The first algorithm, which we call Incremental Latent Confounder search with VBEM (ILC-V), is described in Algorithm 1. It takes a PAG input (Step 1) and uses the ZML algorithm available in R \citep{malinsky} to enumerate all Markov equivalent MAGs of that PAG (Step 3). It then further constructs DAGs for each MAG, starting from the MAGs that contain the minimum number of bi-directed edges (Step 4). Each latent confounder modelled at Step 4 is assumed to be binary, and the optimal DAG is the one that maximises p-ELBO using the VBEM algorithm made available as a Java library by Rodriguez-Sanchez \citeyearpar{mpc}.
\par At Step 5, Algorithm 1 calls Algorithm 1b to determine the optimal number of states for each latent confounder. This is achieved by iterating over each latent confounder present in the highest scoring DAG determined at Step 4, and greedily increasing the number of states by one at a time, for each latent confounder. Algorithm 1b returns a DAG that contains the optimal number of states for each latent confounder, or the maximum number of states $S$ if the objective score continues to increase with the number of states. To improve computational complexity, the objective score p-ELBO is applied to a subgraph $G_S$ that contains the auxiliary latent confounders and their children, since the conditional distributions of the remaining nodes remain unchanged in the BN. The final Step 6 generates the final DAG BN and revises the p-ELBO score.
\begin{center}
\includegraphics[scale=0.64]{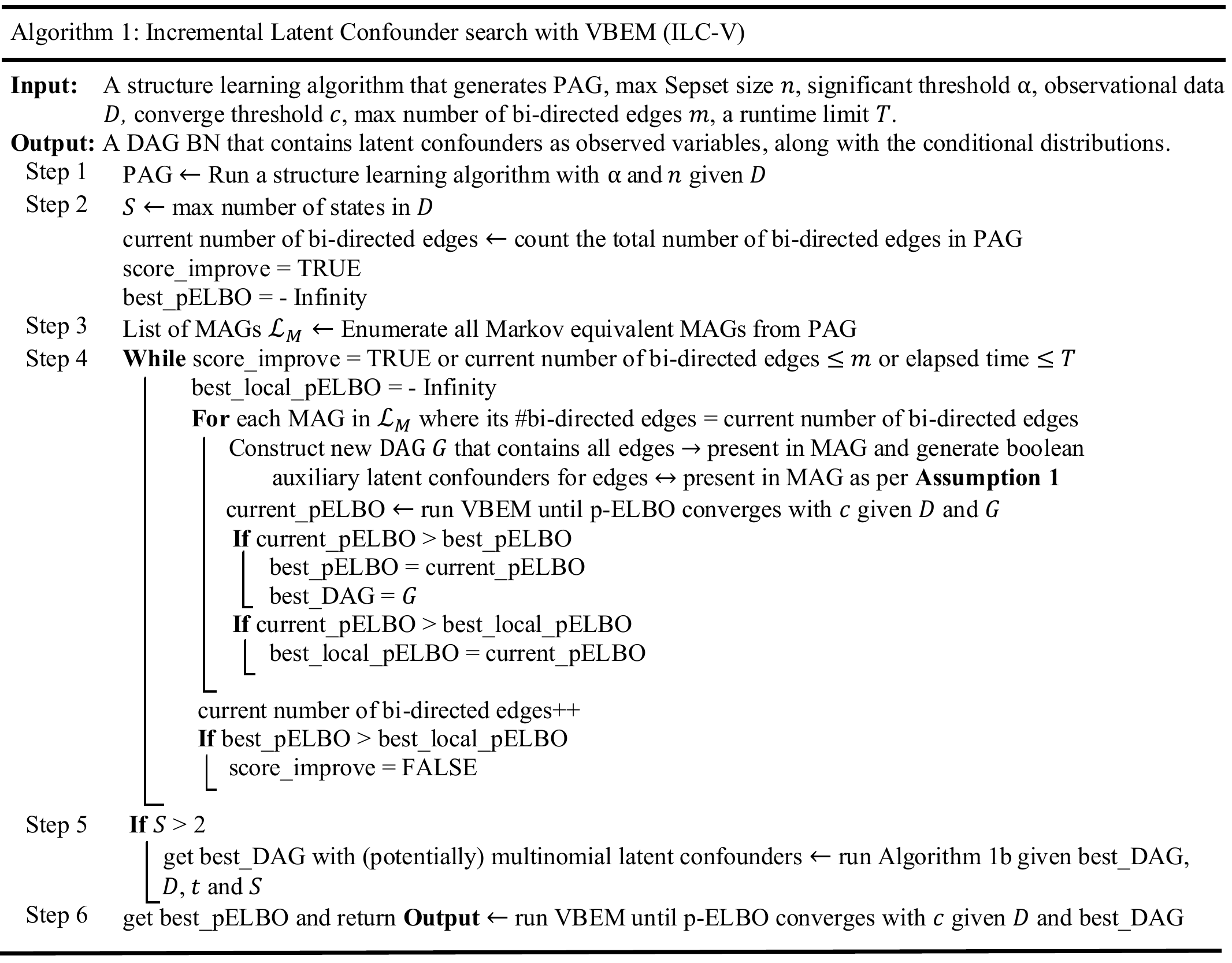}
\end{center} 
\begin{center}
\includegraphics[scale=0.495]{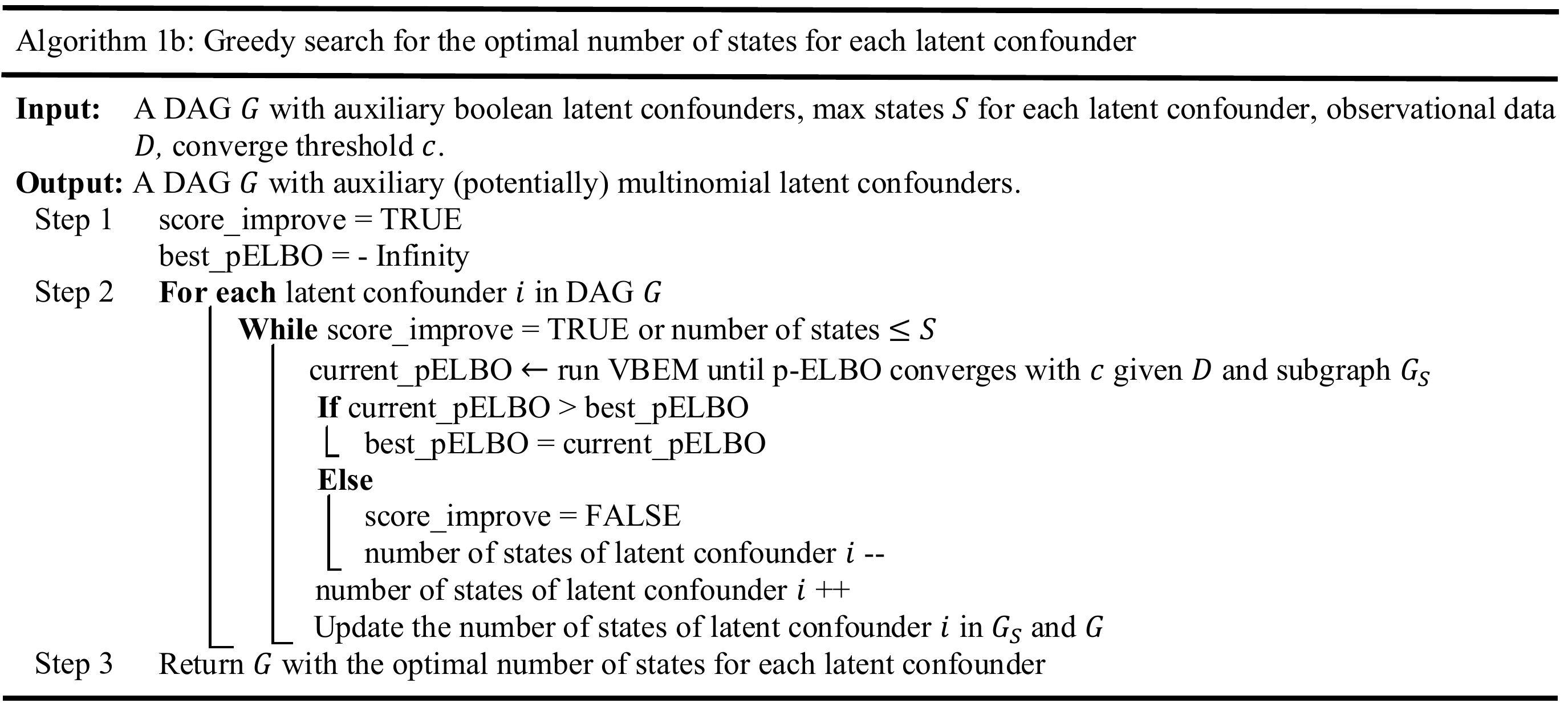}
\end{center} 
\subsection{Alg 2: Hill-Climbing Latent Confounder search with VBEM (HCLC-V)}
\par Because ILC-V (Algorithm 1) is computationally expensive, as we later show in Section 5, one might be interested in a computationally efficient version that minimally decreases the objective score of Algorithm 1. A problem with ILC-V is that when the input PAG contains a high number of invariant edges o—o or o$\rightarrow$, enumerating all possible MAGs can quickly cause memory allocation problems. To address this, we introduce a modified version of ILC-V, which we call Hill-Climbing Latent Confounder search with VBEM (HCLC-V), that skips Markov equivalence checks. This means that HCLC-V no longer needs to check the CIs for each DAG visited, and this saves enormous computational time. Instead, HCLC-V iterates over possible edge orientations as described in Step 4 of Algorithm 2, and continues to follow the incremental search strategy of ILC-V in terms of the number of bi-directed edges. Moreover, a list of the best-found latent confounders from one iteration is carried forward to the next iteration (see Steps 3 and 4 in Algorithm 2). Lastly, since HCLC-V relies on hill-climbing search, it stops exploration when a local maximum is reached.
\begin{center}
\includegraphics[scale=0.65]{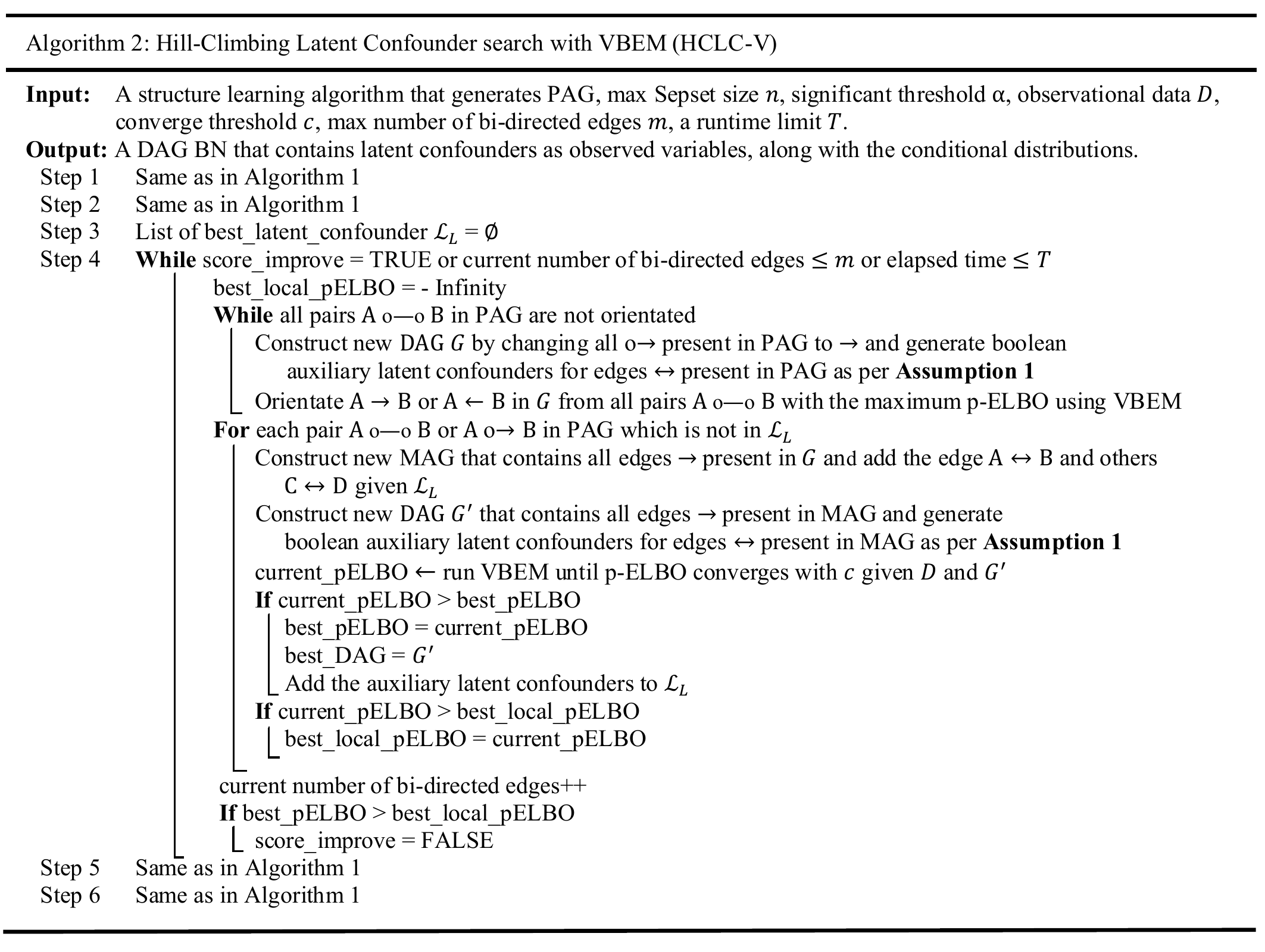}
\end{center} 
\section{Case studies and evaluation setup}
The experimental setup involves four real-world BNs taken from the Bayesys repository \citep{bayesys}, described in Table 1. We generated synthetic data of 1k and 10k samples for each network using the bnlearn R package \citep{bnlearn}. One data set is created for each latent confounder listed in Table 1. This process was applied to both sample sizes, and led to a total of 22 data sets.
\par We have used the constraint-based FCI and the hybrid GFCI algorithms to generate PAGs to be provided as input to ILC-V and HCLC-V. This produced four different result-combinations, which we refer to as ILC-V\mysubscript{FCI}, HCLC-V\mysubscript{FCI}, ILC-V\mysubscript{GFCI} and HCLC-V\mysubscript{GFCI} in Section 5. The GFCI algorithm was tested using the Tetrad-based rcausal R package \citep{rcausal}, and the FCI algorithm was tested using the pcalg R package \citep{pcalg}. Regarding the hyperparameters of FCI and GFCI, we set the G-square significance threshold to $\alpha$=0.05 and the Sepset size to $n$=-1 for unlimited size of conditioning sets. For ILC-V and HCLC-V, we set the maximum number of bi-directed edges to $m$=4 to enable us to carry out experiments within reasonable runtime, and the convergence threshold of VBEM to $c$=0.01.
\begin{center}
\includegraphics[scale=0.65]{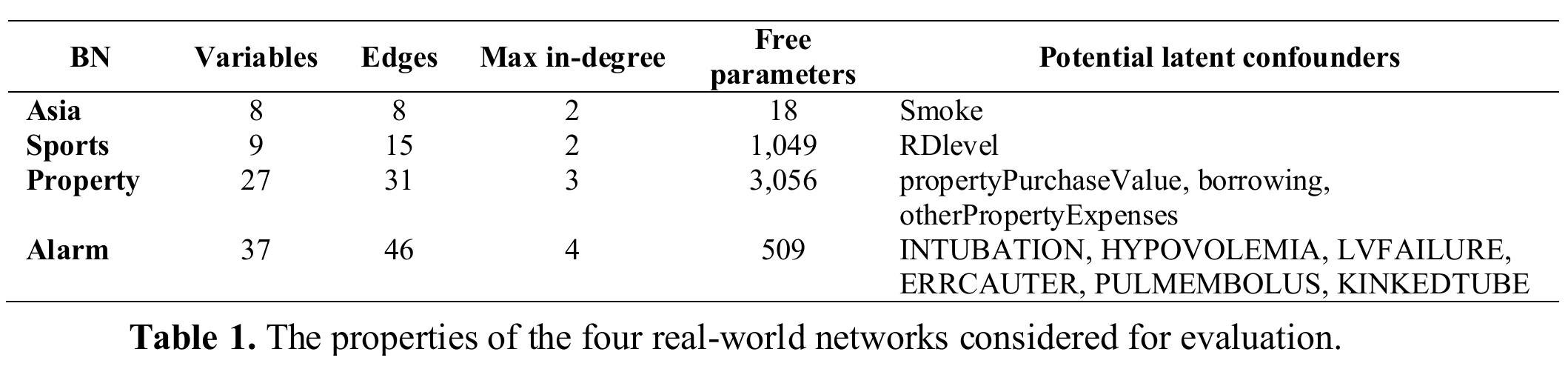}
\end{center}
\par We assess the accuracy of ILC-V and HCLC-V in terms of the objective score p-ELBO and learning runtime, with reference to those obtained by the GLSL and CIL algorithms discussed in subsection 2.4. GLSL and CIL are tested using the Java library by Rodriguez-Sanchez \citeyearpar{mpc} with $mi$=10 regarding the number of pairs of variables to be considered with the highest MI, and with $maxNumberParents\_latent$=-1 for GLSL to assume no parents for density estimation of latent confounders to enable us to carry out experiments within reasonable runtime.
\par We impose a runtime limit of 12 hours for each experiment and set hyperparameter $T$ to 12 hours for both ILC-V and HCLC-V, to ensure that they return a result within the 12-hour runtime limit. Experiments by the other algorithms that do not complete learning within the specified runtime limit are denoted as “Timeout”. All experiments are based on 8GB of RAM. The experiments involving the Asia, Sports and Property networks were carried out on the Intel Core i5-8250 CPU at 1.80 GHz, whereas the experiments involving the Alarm network on the M1 CPU at 3.2 GHz.
\section{Empirical results}
\subsection{The difference in search space explored by ILC-V and HCLC-V}
This subsection investigates the difference in search space explored between the two proposed algorithms, ILC-V and HCLC-V. The comparison assumes that the PAG inputs are produced by GFCI, and relies on Step 4 (which represents the main difference between the two algorithms) where the latent confounders are assumed to be binary.
\par Figure 2 presents the results based on the Property network (27 nodes) for both sample sizes 1k and 10k. Figure 2a shows that ILC-V\mysubscript{GFCI} produces a slightly higher p-ELBO score than HCLC-V\mysubscript{GFCI}, but that ILC-V\mysubscript{GFCI} achieved that by exploring considerably more search space than HCLC-V\mysubscript{GFCI}; i.e., visited a total of 170 DAGs versus 20 DAGs. The charts depict different colours to illustrate how the two algorithms differ at visiting DAGs derived from MAGs that contain increasing numbers of bi-directed edges. Specifically, Figure 2a shows that ILC-V\mysubscript{GFCI} visited all DAGs derived from MAGs containing up to three bi-directed edges, whereas HCLC-V\mysubscript{GFCI} ended at a local maximum while visiting DAGs derived from MAGs containing up to two bi-directed edges.
\par Figure 2b, on the other hand, shows that the higher sample size helped ILC-V\mysubscript{GFCI} to both find a higher objective score and complete learning faster than HCLC-V\mysubscript{GFCI}. This is because ILC-V\mysubscript{GFCI} found no DAG derived from MAGs containing two bi-directed edges to have a higher score than the highest scoring DAG derived from MAGs containing one bi-directed edge, which caused ILC-V\mysubscript{GFCI} to skip MAGs containing three bi-directed edges. On the other hand, HCLC-V\mysubscript{GFCI} ended up visiting a higher number of DAGs, but note this does not necessarily imply that the algorithm was slower; i.e., recall that HCLC-V skips checking for Markov equivalence between graphs.

\includegraphics[scale=0.62]{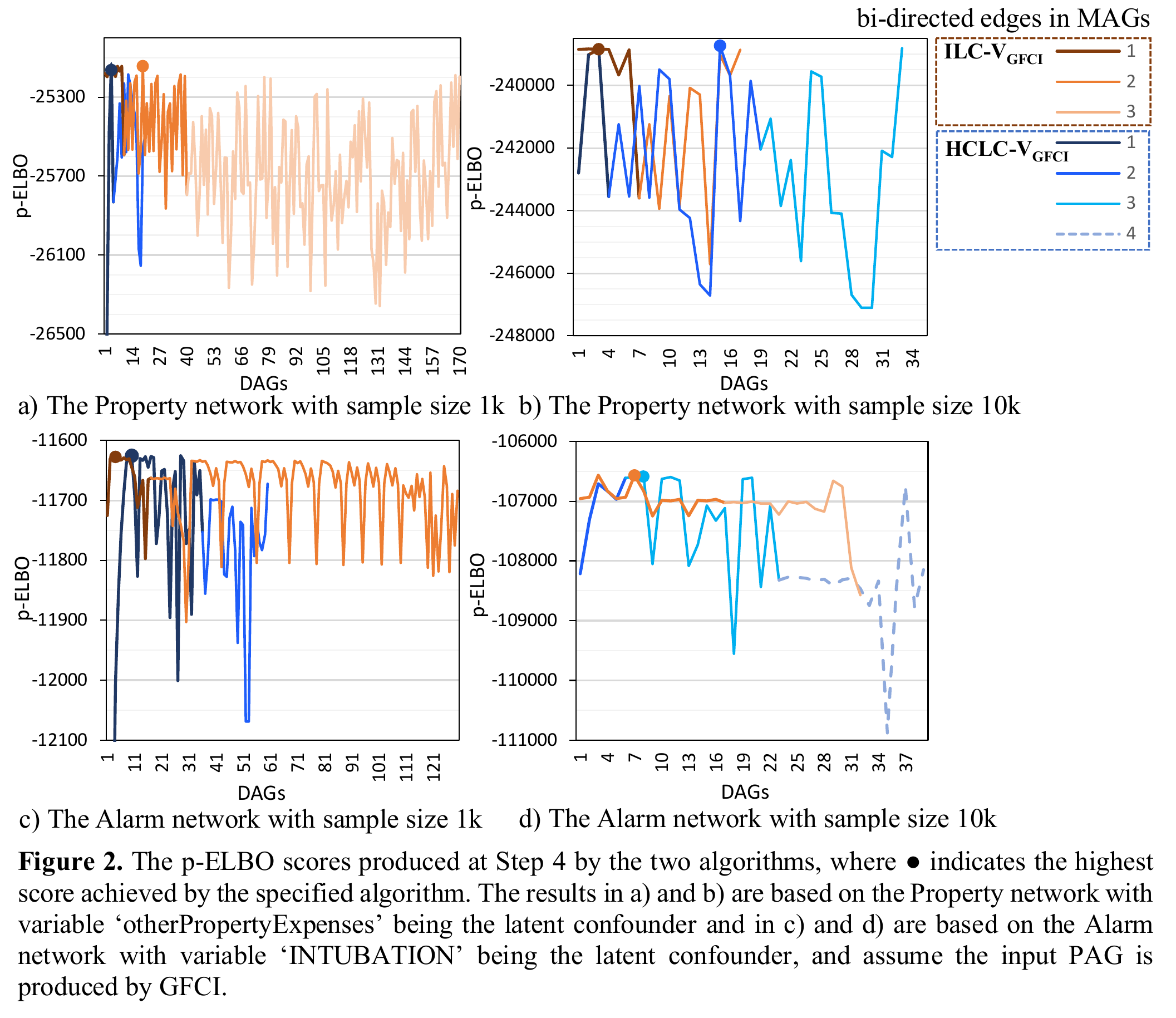}

\par Figure 2c and 2d repeat the analysis of Figure 2a and 2b with application to the Alarm network (37 nodes), and show that the results are consistent with those produced for the Property network. The only difference here is that, at 10k sample size, the p-ELBO score of HCLC-V\mysubscript{GFCI} matched that of the generally slower ILC-V\mysubscript{GFCI}.
\subsection{Performance of ILC-V and HCLC-V relative to other algorithms}
We compare the results produced by ILC-V and HCLC-V to those produced by the CIL and GLSL algorithms described in subsection 2.4 which, to the best of our knowledge, are the two algorithms that are most relevant to this work, which involves both the discovery and density estimation of latent confounders.
\par Table 2 presents the p-ELBO score for each algorithm and data set combination, plus the p-ELBO scores of the true DAGs, for both sample sizes 1k and 10k. The average ranks show that ILC-V\mysubscript{GFCI} performs best in terms of maximising the p-ELBO score across both sample sizes, followed by HCLC-V\mysubscript{GFCI}. CIL algorithm is found to be the worst performing algorithm at sample size 10k, whereas GLSL mostly outperforms both ILC-V\mysubscript{FCI} and HCLC-V\mysubscript{FCI}, but not ILC-V\mysubscript{GFCI} and HCLC-V\mysubscript{GFCI}. This means that ILC-V and HCLC-V benefit from the PAG input of GFCI, and suggests that the hybrid learning GFCI might be better than FCI at recovering PAGs; an observation consistent with previous studies \citep{Largescale}. Note that while the true DAG will not always have the highest p-ELBO score, the highest scores produced by the algorithms tend to be very close to those of the true DAG, and this helps to validate the results.	
\begin{center}
\includegraphics[scale=0.52]{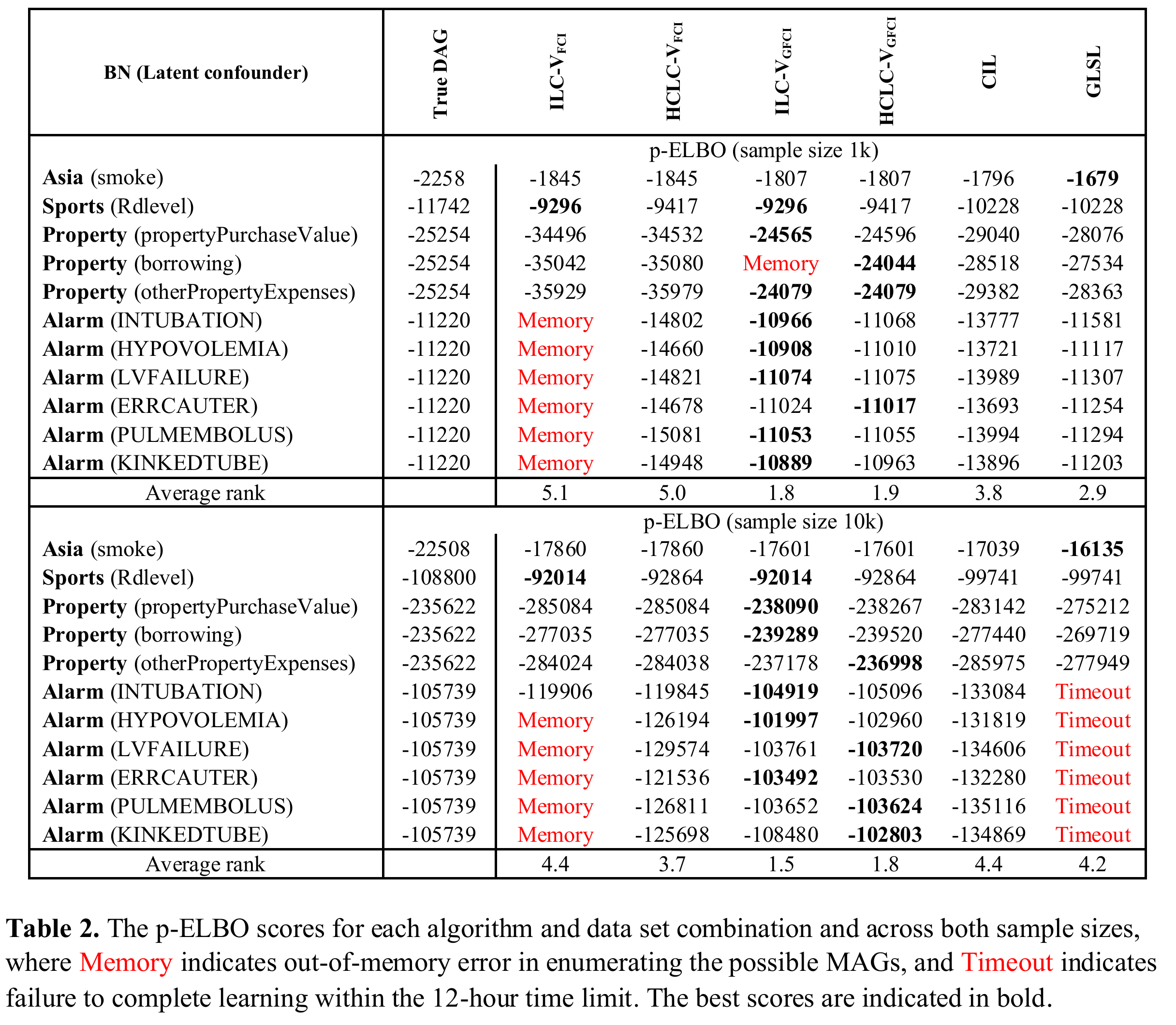}
\end{center} 
\par While ILC performs best in general, it does not scale well with the size of the network. As shown in Table 2, ILC-V returns an out-of-memory error (for 8GB RAM) for most experiments with Alarm, specifically when paired with FCI, caused by the large number of possible MAGs derived from the input PAG. The cumulative runtime across all 10k sample sizes was 14, 34, 46 and 88 hours for CIL, HCLC-V\mysubscript{GFCI}, ILC-V\mysubscript{GFCI} and GLSL respectively, with a similar trend observed across 1k sample sizes. On average, HCLC-V is found to be 1.4 times faster than ILC-V, which in turn is found to be 1.6 times slower than CIL and 4.5 times faster than GLSL which failed to complete the Alarm network experiments at 10k sample size; suggesting that its computational efficiency might not scale well with sample size.
\section{Conclusions}
This work investigated two novel algorithms that can be used for both discovery and density estimation of latent confounders in BN structure learning from discrete observational data. The first algorithm (ILC-V) aims to maximise model selection accuracy by exploring sets of Markov equivalent MAGs, starting from the set of MAGs that contain the lowest number of bi-directed edges and, while the objective score increases with each set, moving to sets of MAGs with increasing numbers of bi-directed edges. The second algorithm (HCLC-V) aims to balance accuracy relative to computational efficiency by employing hill-climbing over the MAG space, enabling application to larger networks.
\par Both algorithms require a PAG to be provided as an input, which means that the proposed algorithms need to be paired with a structure learning algorithm that recovers ancestral graphs. Because the input PAG will typically indicate multiple possible latent confounders, the ILC-V and HCLC-V algorithms use p-ELBO as the objective function to determine the number as well as the position of the latent confounders, thereby contributing to the discovery process, in addition to density estimation, of latent confounders. 
\par The two proposed algorithms are evaluated relative to two recent and relevant implementations that also optimise for p-ELBO. The empirical results show meaningful improvements in maximising the objective score, and in some ways in reducing time complexity; although the latter remains a major issue. Two important limitations are that a) both algorithms rely on a PAG input to be provided by some other structure learning algorithm, and b) the results are based on experiments that assume a single latent confounder only, which was necessary to ensure that most experiments complete within the 12-hour runtime limit.
\vskip 0.2in
\bibliography{references}
\end{document}